\documentclass[letterpaper, 10 pt, conference]{ieeeconf}  % Comment this line out if you need a4paper

\IEEEoverridecommandlockouts                              % This command is only needed if 
\usepackage{graphicx} % for pdf, bitmapped graphics files
\usepackage{times} % assumes new font selection scheme installed
\usepackage{amsmath} % assumes amsmath package installed
\usepackage{amssymb}  % assumes amsmath package installed
\usepackage{url}
\usepackage{cite}
\usepackage{enumerate}
\usepackage{bm}
\usepackage{cases}
\usepackage{mathtools}
\usepackage{algorithmicx}
\usepackage{algorithm}
\usepackage{algpseudocode}% http://ctan.org/pkg/algorithmicx
\usepackage{booktabs, ragged2e}
\usepackage{tabularx}
\usepackage{graphicx}
\usepackage{multirow}
\usepackage[para]{threeparttable}
\usepackage[table,xcdraw]{xcolor}

\makeatletter
\let\NAT@parse\undefined
\makeatother
\usepackage[hidelinks]{hyperref} %hide the borders of the cross-references!

\DeclareMathOperator*{\argmin}{argmin}%

\newcolumntype{C}{>{\Centering\arraybackslash}X}
\newcolumntype{L}{>{\raggedright\arraybackslash}X}
\newcolumntype{R}{>{\raggedleft\arraybackslash}X}
\begin{document}
\title{\LARGE \bf
        \textit{Velocity Field}: An Informative Traveling Cost Representation for Trajectory Planning

        \author{Ren Xin$^{1,2}$, Jie Cheng$^2$, Sheng Wang$^{1,2}$ and Ming Liu$^{1,2,3}$}% <-this % stops a space

        \thanks{
        This work was supported by Guangdong Basic and Applied Basic Research Foundation, under project 2021B1515120032, National Natural Science Foundation of China under Grants 62333017, and Project of Hetao Shenzhen-Hong Kong Science and Technology Innovation Cooperation Zone(HZQB-KCZYB-2020083), awarded to Prof. Ming Liu. (Corresponding author: Ming Liu.)
        }
        \thanks{
        $^1$The Hong Kong University of Science and Technology (Guangzhou), Nansha, Guangzhou, 511400, Guangdong, China. $^2$The Hong Kong University of Science and Technology, Hong Kong SAR, China. $^3$HKUST Shenzhen-Hong Kong Collaborative Innovation Research Institute, Futian, Shenzhen, China. (Email: \texttt{\{rxin, jchengai, swangei\}@connect.ust.hk}, \texttt{eelium@ust.hk}).
        }
        %\thanks{$^3$Open-source implementation will be released at \url{https://github.com/jchengai/gpir}.}%
        % , Jianhao Jiao jjiao
}
\maketitle
\thispagestyle{empty}
\pagestyle{empty}

%%%%%%%%%%%%%%%%%%%%%%%%%%%%%%%%%%%%%%%%%%%%%%%%%%%%%%%%%%%%%%%%%%%%%%%%%%%%%%%%
% 对于自动驾驶来说，规划的可解释性是至关重要的。一个具有强可解释性的规划器可以缓解乘坐人员的焦虑，帮助研发人员解决问题以促使无人系统普及。
% 本文提出了一种隐式的静态地图表征和可视化方法以解决过去基于格点的方法中地图分辨率和预测范围之间的矛盾以及推理时间长的问题，并且提高了直接基于矢量图的地图信息归纳能力。
% 相关的开闭环实验证明该方法在地图的表达能力和规划系统的可靠性中分别有x%， x%的提升。

\begin{abstract}
        % \bt{
        Trajectory planning involves generating a series of space points to be followed in the near future. However, due to the complex and uncertain nature of the driving environment, it is impractical for autonomous vehicles~(AVs) to exhaustively design planning rules for optimizing future trajectories. To address this issue, we propose a local map representation method called \textit{Velocity Field}. This approach provides heading and velocity priors for trajectory planning tasks, simplifying the planning process in complex urban driving scenarios. 
        The heading and velocity priors can be learned from demonstrations of human drivers using our proposed loss functions. 
        Additionally, we developed an iterative sampling-based planner to train and compare the differences between local map representation methods.
        We investigated local map representation forms for planning performance on a real-world dataset. 
        Compared to learned rasterized cost maps, our method demonstrated greater reliability and computational efficiency.
        
        The video can be found at \href{https://youtu.be/5LuyGN58CD0}{\color{blue}https://youtu.be/5LuyGN58CD0}.

% \begin{keywords}
%         Autonomous vehicle navigation, learning-based planning, motion and path planning.
% \end{keywords}
\end{abstract}

%%%%%%%%%%%%%%%%%%%%%%%%%%%%%%%%%%%%%%%%%%%%%%%%%%%%%%%%%%%%%%%%%%%%%%%%%%%%%%%%

\section{Introduction}

\begin{figure}[ht]
  \centering
  % \subfloat[Single modal planner]{
  \includegraphics[width=3.3in]{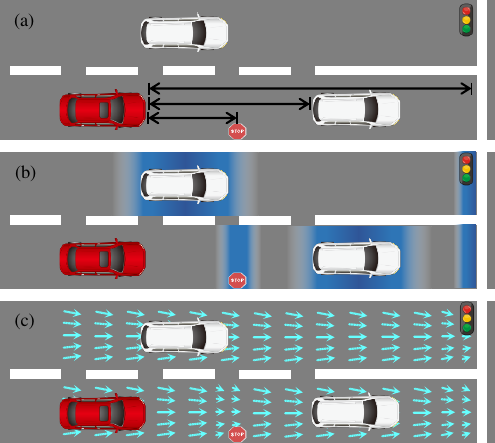}
  % }
  % \hfill
  % \subfloat[Multi-policy planner (ours)]{\includegraphics[width=1.65in]{images/page1_right.pdf} \label{fig:multi-policy}}
  \caption{
    This figure illustrates three methods for calculating the cost of travel at a given time step. The red car represents the ego vehicle, the solid white line on the far right depicts the stop line for the traffic light, and the traffic lights in each scenario are in the red light state.
    (a) distances between the ego vehicle and other vehicles and traffic signals are measured. Manually defined rules are then implemented to determine the car-following speed and lane change maneuvers.
    (b) represents a class of map representations that provide the driving cost for each point in the driving context. It guides the vehicle trajectory through as many low-driving cost areas as possible.
    (c) uses the velocity vector to guide the vehicle's driving trajectory. For example, it instructs vehicles to slow down and stop before the stop line around stop signs and to drive at the same speed as other cars.
    % TODO:%头图大概可以放三张图片？分别是单纯的distance measurement，基于逆卷积的cost map和基于attention的速度场?
    % Planning under urban scenarios is a challenging task because it needs to consider the perception, prediction, and control uncertainties together with complex road environments. 
    % Our method reflects all perception information in a risk map, which is a middle-ware of sensing and planning.
    % The purple rectangle represents the ego vehicle, and the red blocks are other agents. 
    % The transparency of \textbf{orange-yellow} fields represents the occupancy probability of other traffic participants. 
    % The \textbf{white-yellow-orange} points represent the risk value of sampled positions. 
    % Sample points with higher risk values are redder.
    % (a) For the single modal planner, it is generally difficult for the model to learn behaviors other than lane keeping, as it is the dominant behavior in the dataset. 
    % (b) Our proposed multi-policy planner, which forces the model to explicitly think about multiple policies and choose the most suitable policy given the current driving context.
  }
  \label{fig:cover_img}
\end{figure}

% 中文版
% 路径规划是自动驾驶问题中的重要模块。它为车辆提供一条可跟踪的安全轨迹，以使智能体尽快到达目标位置。
% 现存的规划方法主要分为两类，传统方法和基于模仿学习的方法。
% 传统方法通过人工设计策略避免碰撞场景中各类可感知的物体。而人工设计的策略通常只能适用于某一类场景。这导致传统方法的泛化能力不强，只适用于有限的场景中。
% 基于模仿学习的方法本质上是学习了轨迹与场景之间的映射关系。这种方法免于手工设计策略从而更适合应用于更广泛的场景。但是大量的模型参数导致其可解释性不强。
% NMP提出了一种新方法，它能够结合前两种方法的优势，从而同时提高规划模块的泛化性和可解释性。深度神经网络被用于将感知信息转化为场景中每个时空点上的行驶代价，并且通过采样的方式选择总代价最低的轨迹。我们复现了这种方法，由于其cost map生成时基于卷积神经网络的，需要较大计算量，无法满足实时性要求。同时我们发现其在高速状态下会出现地图范围不足的问题，而过大的grid size又会造成规划精度的下降。此外，大范围的grid中有很大一部分在车后的或不可能行驶到的区域位置也被预测了，浪费了大量计算资源。
% 因此该方法面临着计算时间，地图范围和grid size三者之间的trade off。
% 此外，我们认为单纯的cost value提供的信息是基于位置的，没有考虑更高阶的运动信息。比如车辆开向路边和沿着路边开的cost应该是不同的，并且cost map无法表达道路限速信息。
% 因此，对于trade off的问题，我们提出了隐式地图的概念。隐式地图基于attention模块实现。场景信息被深度神经网络编码后作为键值，轨迹采样位置则被编码为query来查询该位置的行驶代价。这种形式避免了显式得将场景编码解码为二维栅格地图。并且避免了输出无关位置的行驶代价。因此具有更高的计算效率，且不存在grid range 和 grid size的trade off。
% 此外，类似于occupancy map和occupancy flow的关系，我们将原来的静态的cost value改为了速度向量先验来引入高阶运动信息，通过计算轨迹速度与地图先验速度的差值来估计行驶代价。
% 我们的方法在记录了现实场景的数据集中进行了验证，结果证明我们的方法在安全性和类人性上均好于其他方法。

Trajectory planning is an essential component of autonomous driving systems\cite{urban_driver}, as it makes the vehicle follow a target path to the intended destination with the promise of efficiency and safety.
Existing planning methods can be classified into two categories: rule-based methods and learning-based methods.
% Existing planning methods are mainly divided into two categories, traditional methods, and methods based on Imitation Learning~(IL).

% Traditional methods rely on distance measurement and artificially designed strategies to avoid collisions with perceivable objects in the driving context, limiting their generalization ability. 
% IL-based methods, on the other hand, learn the mapping relationship between trajectory and driving context, offering a wider range of applications. 
% However, the massive parameters of neural networks lead to poor interpretability.
Rule-based methods rely on manually designed rules to avoid collisions with objects in the driving context, such as distance measurement and velocity strategy based on distance, as shown in Fig. \ref{fig:cover_img}~(a). 
However, these rules are often limited to specific scenarios, thus restricting their generalization ability.
Learning-based methods, particularly those based on imitation learning, learn the mapping between trajectory and driving context, providing a broader range of applications. 
Nevertheless, directly mapping driving context to planning trajectory is often considered poorly interpretable~\cite{NMP}.

The Neural Motion Planner~(NMP)\cite{NMP} introduced a novel approach that integrates the benefits of rule-based and learning-based methods, thereby improving the generalization and interpretability of the planning module.
Specifically, this method employs a deep neural network to convert perceptual information into driving cost at each space-time point in the scenario~(Fig. \ref{fig:cover_img}~(b)), and selects the sampled trajectory with the lowest total cost.
However, the cost map generated by convolutional neural networks imposes a significant computational burden and fails to meet real-time requirements.
Additionally, planning in high-speed scenarios necessitates a larger map range and an excessively large grid size results in reduced planning accuracy. 
At the same time, a considerable portion of the large-scale grid behind the car or in unreachable areas is also predicted, resulting in a waste of computing resources.
Consequently, this method faces a trade-off between computation time, map range, and grid size.
Moreover, the pure cost value only provides information about spatiotemporal location, without considering higher-order motion information. For instance, the cost of driving to the roadside and driving along the roadside should differ, and the cost map cannot convey information about the road's speed limit.

To address these challenges, we propose the \textit{Velocity Fields}~(VF), which comprises two novel key designs.
First, to overcome the computational tradeoff, we propose the concept of implicit maps implemented based on the attention mechanism\cite{attention}. 
 In our method, the driving context information is encoded as key-value pairs, while the trajectory sampling position is encoded as a query to obtain the driving cost at that position. 
This approach avoids explicitly decoding the latent variable into a rasterized map and outputting the cost value of irrelevant positions, resulting in higher computational efficiency and no trade-off between grid range and size. 
Second, similar to the relationship between the occupancy map\cite{OCC_Map} and occupancy flow\cite{occflow}, we replace the original cost values with the velocity vectors~(Fig. \ref{fig:cover_img} (c)), which introduces higher-order motion information. The traveling cost is estimated by computing the difference between the trajectory velocity and the velocity priors.
We validate our method on a dataset that records real-world scenarios.
% and the results show that our approach outperforms other methods in terms of security and human similarity.

Our contributions are mainly three folds:
\begin{enumerate}

    % \item In the framework based on first prediction and then planning, a more efficient and flexible expression of static obstacles is plugged in so that it can plan trajectories more similar to human drivers.
    \item We explore a novel interpretable vectorized driving local map representation method, termed \textit{Velocity Field}~(VF), which boosts the planning performance in a straightforward and efficient manner.
    % which provides second-order planning priors in a straightforward and efficient manner.
    
    % \item The planning results are expressed in the form of an implicit velocity field, which improves the computational efficiency and reduces the space complexity of map visualization. 
    \item We develop an efficient iterative trajectory optimizer that is seamlessly compatible with the proposed map representation method, enabling both the training and inference process.
    
    \item We deploy \textit{Velocity Field} and iterative optimization-based planners in the recorded scenarios from the real world in closed-loop form, demonstrating the human similarity and safety improvements achieved by our proposed method.
\end{enumerate}
\section{Related Work}

% There are extensive literature on behavior planning from an interactive multi-agent perspective, and many of them are formulated as partially observable Markov decision processes (POMDPs)
% POMDP-based method is the mainstream 

% \subsection{Multi-modal Motion Forecasting}
% State-of-the-art multi-modal motion forecasting

% 
\subsection{Environment Feature Extraction}

Due to advancements in feature extraction using Deep Neural Networks (DNN), numerous learning-based planning approaches have been proposed \cite{CoverNet, Chauffeurnet}. 
These approaches take rasterized maps and vehicle history trajectories as input and output different hypotheses assigning probabilities to each trajectory. 
Some alternative methods \cite{vectornet, PGP} encode contextual information in vector and graph formats, which have proven to be more powerful due to their enhanced efficiency in context representation. 
Furthermore, these methods employ attention mechanisms or graph convolution techniques to aggregate information and learn latent interaction models, enabling the planner to focus more on the most relevant information.
Jie \textit{et al.} \cite{cj2022mpnp} proved that emphasizing domain knowledge by explicitly broadcasting map elements can improve performance on related aspects like lane selection. 
These feature extraction and processing methods inspired the encoder design of our approach.

% NERF\cite{mildenhall2020nerf} and DETR\cite{carion2020end} rethink the role of neural networks in 3D modeling and object detection problems. Both of them use an inquiry framework. NERF considers perspective information as input and output rendered images directly. DETR deprecate traditional complex anchor-based method, learned object category embedding are token inquiry of their positions.

\subsection{Representation of Local Map Information}
The local map differs from the global map as it expresses scene information in the presence of time-dependent traffic participants and rules. 
% existing global routing information. 
% The role of the local map is mainly to carry out local collision avoidance when there is a general direction of travel or a reference route. 
In the past few decades, the concept of occupancy grid maps \cite{OCC_Map} and configuration space had been widely used to empower planning algorithms of autonomous driving. 
However, this method only enables the agent to have basic obstacle avoidance capabilities and cannot meet the requirements of following traffic regulations on urban roads. 
HD maps \cite{hdmap} offer semantic information that is essential for urban planning, including lanes, stop lines, and traffic signs. They have been widely used in rule-based planning methods. 
Our method also makes good use of the semantic information and integrated them with learning-based methods by rule check.  
Learning-based research tended to improve their reliability by constraining resulting trajectories using occupancy maps \cite{learn2predict, unlikelihood} or by calculating distances from the reference line \cite{MP3}. 
% Rasterized probabilistic occupancy maps \cite{HOME} can be adopted to improve the rationality of target position distribution in both planning and prediction tasks. 
NMP \cite{NMP} explored the use of neural networks to generate spatiotemporal cost maps using deconvolution modules. 
% The neural network training utilizes recorded expert trajectories as positive samples while constructing negative samples based on distance and traffic rules. 
However, cost map-based methods can only offer first-order planning information which is lack of direction and speed guidance. 
%still have ample room for improvement in terms of overall planning performance.

Vector Field is an informative local map representation as it can offer more robust guidance for path following \cite{Nelson2007VFPF, lawrence2008lyapunov}. 
However, it has only been applied to the control problem of tracking reference lines. In contrast, we combined it with deep neural networks to deal with complex road conditions.
Recently, the implicit occupancy flow field\cite{agro2023implicit} is proposed to predict occupied space regarding time by cross attention which is similar to ours. 
They are primarily concerned with motion prediction and its accuracy, whereas we are concerned with the construction of planning guidance based on scenario information.

% put in related works?
\subsection{Inverse Optimal Control}
% There are rich literatures on ML-based planner for Self-driving.
% There are several main branches of Learning from experts, Inverse Optimal Control(IOC), Behavior Clone(BC), and Inverse Reinforcement Learning(IRL). 
Inverse Optimal Control(IOC) is to learn the parameterized strategy to deal with objects in traffic scenes during driving from demonstration data\cite{yaman2013survey}.
Levine \textit{et al.} \cite{levine_continuous} adopted this method to regress the parameters of the planning strategy. % levine_nonlinear
%represent the reward function as a Gaussian Process(GP) that maps measurements from feature values to rewards by IOC methods.%, which inspired me to use exponential to uniform measurements.
% while inverse IRL is to recover a cost function based on a deep neural network in relatively complex systems\cite{review_ioc}. 
% In 1826 Abel first mathematically studied the inverse mechanical problems for finding the curve of an unknown path .
% Discrete-time IOC scheme for trajectory tracking of non-linear systems is developed by \cite{discreteIOC}.
% Imitation Learning
% Both BC and IRL can be categorized as Imitation Learning. 
% Thanks to the development of parallel programming, large-scale neural networks are becoming more and more widely used. 
% BC directly establishes a mapping from inputs to action, while IRL is trying to find a reward function for behavior decisions.
Previous works \cite{NMP, MP3} have integrated end-to-end learning with a sampling-based planner, achieved either by estimating a cost volume or predicting diverse future trajectories of other agents. % lookout
However, their trajectory samples use parameterized curves or clustered trajectory logs, resulting in sparse data that cannot meet the needs of the planner. 
Furthermore, the reproducibility of their results is limited due to reliance on private datasets, which restricts further validation and application of the methods. 
DIPP \cite{DIPP} is the first to combine numerical optimization approaches with regressive model parameters by ensuring the trajectory optimization process is differentiable, achieving state-of-the-art planning and prediction accuracy.

\begin{figure*}[ht]
  \centering
  \includegraphics[width=6.6in]{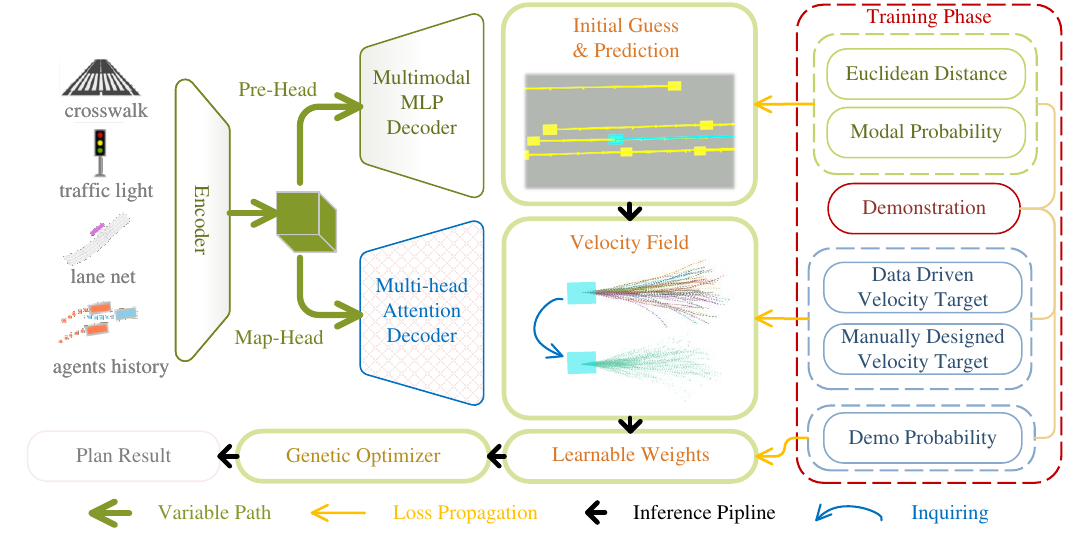}
  
  \caption{The diagram presents our proposed framework for velocity field generation and planning modules, along with their training process. 
  The \textcolor[RGB]{62, 179, 61}{\textbf{green arrows}} represent the transmission of latent variables. 
  Reference lines, map elements, ego and agent motion, and geometry are encoded using MLP/GRU modules. 
  Global attention is employed to model the interaction between objects. 
  The multi-modal MLP decoder generates initial guesses for the ego vehicle and predictions for other agents. 
  The \textcolor[RGB]{255, 192, 0}{\textbf{orange arrows}} indicate the relationship between loss and each module: 
  expert demonstrations provide supervisory signals to model parameters through loss functions, such as distance measures, selection probabilities, and velocity targets. 
  The \textbf{black arrows} represent the planning process: the iterative trajectory optimizer samples around the initial guess; calls the \textcolor[RGB]{7, 86, 245}{inquiry function} to obtain speed information at each point of \textcolor[RGB]{0, 255, 255}{ego vehicle} trajectory candidates; calculates the cost of each track through weighted summation; select the best n trajectories for the next iteration; and after a certain number of iterations; selects the one with minimum cost as the final output trajectory.
    % \textbf{Planning:} A sampling-based planner is adopted, and all the sampled spatial-temporal points are with distance information to representative represent driving contexts.
    % The global encoded latent variable in the prediction module generates parameters mapping distance to risk value space.
    % Risk values are concatenated with collision risk. 
    % The trajectory with the minimum risk is selected as the planning result.
    % \textbf{Training:} A two-stage supervised learning process is adopted. The predictor and planner are trained with listed metrics respectively.
  }
  \label{fig:overview}
\end{figure*}
\section{Methodology}
\label{sec:method}
% The visio graph put here

\subsection{Problem Formulation}
% This task is to generate a set of future trajectory points of the ego vehicle $\mathbf{P} = \left[\bm{p}^1, \dots, \bm{p}^T \right]$ in a short time period $T$ subject to history tracking data of other agents $h_a$ and ego vehicle $h_e$, map information of current time $\bm{\mathcal{M}}$ include reference lane $m_{ref}$, traffic light status $m_{tl}$, and static obstacles $m_{ob}$. 
% The driving context is denoted as $\mathcal{S}  = (\bm{h_a},\bm{h_e},\bm{\mathcal{M}})$ and our model is to find a mapping relationship $\mathbf{P} = \mathcal{F} (\mathcal{S})$.
The objective of this task is to generate a set of future trajectory points for the ego vehicle, denoted as $\mathbf{P} = \left[\bm{p}^1, \dots, \bm{p}^T \right]$, over a short time period $T$, given the historical tracking data of other agents $h_a$ and the ego vehicle $h_e$, as well as the map information of the current time $\bm{\mathcal{M}}$, which includes the reference lane $m_{ref}$, traffic light status $m_{tl}$, and static obstacles $m_{ob}$.
The driving context is represented by $\mathcal{S} = (\bm{h_a},\bm{h_e},\bm{\mathcal{M}})$, and our model aims to establish a mapping from the driving context to trajectory points $\mathbf{P} = \mathcal{F} (\mathcal{S})$. 
The overall method is illustrated in Fig. \ref{fig:overview}.

\subsection{Driving Context Encoding} \label{sec:driving-context}
Similar to Vectornet \cite{vectornet}, we implement a graph neural network to encode vectorized context information for the prediction module. Both the $T$ time step ego-vehicle history $h_e \in \mathbb{R}^{T\times [x,y,yaw,\dot{x},\dot{y}]}$ and $N_a$ surrounding vehicles $h_a \in \mathbb{R}^{N_a \times T \times [x,y,yaw,\dot{x},\dot{y},type]}$ features are concatenated together and embedded by a Gated Recurrent Unit (GRU) to encode the time-dependent information.
Subsequently, the features extracted in a 256-dimensional space are processed by a multi-layer perceptron (MLP). Map elements like piece-wise lanes and crosswalks polyline are initially directly encoded by MLP.
All the 256-dimensional encoded features are concatenated in the first dimension and aggregated by a global multi-head attention network (GAT) to obtain the vehicle-to-vehicle and vehicle-to-context relationships.
The overall scenario context embedding is symbolized as $\mathcal{S}$.

This module generates predicted future trajectories of neighboring agents and an initial planning guess for the ego vehicle. Trajectory replays from the dataset directly serve as supervisory signals by computing the smoothed L1 loss, thereby guiding the trajectory optimization process.

% \subsection{Multi-policy Encoding} \label{sec:multi-policy}
\subsection{Iterative Gaussian Sampling Trajectory Optimizer }
During the training and inference process, we adopted an iterative Gaussian sampling-based planner to enhance sampling efficiency and improve sampling accuracy. The optimization process is similar to STOMP \cite{STOMP}.
In our work, the sampler performs Gaussian sampling regarding control variables including acceleration and steering $(a,s)$, which is obtained by differentiating the initial guess.
It selects the best k samples as new means and resamples until the diagnosed iter limit. 
Three categories of trajectories constitute the samples in the first loop: a control variable probabilistic grid sampler, a lattice state sampler \cite{lattice}, and the initial guess based on vanilla imitation learning. 
The sampling variance for acceleration and steering is controlled by two independent hyperparameters $(\sigma_a, \sigma_s)$ and sampled independently.
% , ensuring optimal trajectory planning.

It is important to note that the disturbance in the first iteration is time-constant, while in the subsequent iterations, it is time-dependent with decreasing variance. The first loop aims to ensure that the most feasible trajectories are explored, while the following iterations optimize the best choice.

\subsection{Implicit Velocity Field}
% DE:TR style presentation image/Velocity field ground truth image
\begin{figure}
  \centering
  % \subfloat[Single modal planner]{
  \includegraphics[width=3.3in]{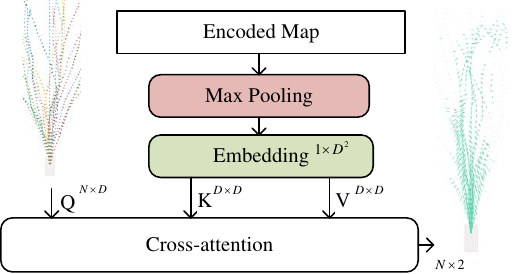}
  % }
  % \hfill
  % \subfloat[Multi-policy planner (ours)]{\includegraphics[width=1.65in]{images/page1_right.pdf} \label{fig:multi-policy}}
  \caption{
    %头图大概可以放三张图片？分别是单纯的distance measurement，基于逆卷积的cost map和基于attention的速度场?
    % Planning under urban scenarios is a challenging task because it needs to consider the perception, prediction, and control uncertainties together with complex road environments. 
    % Our method reflects all perception information in a risk map, which is a middle-ware of sensing and planning.
    % The purple rectangle represents the ego vehicle, and the red blocks are other agents. 
    % The transparency of \textbf{orange-yellow} fields represents the occupancy probability of other traffic participants. 
    % The \textbf{white-yellow-orange} points represent the risk value of sampled positions. 
    % Sample points with higher risk values are redder.
    % (a) For the single modal planner, it is generally difficult for the model to learn behaviors other than lane keeping, as it is the dominant behavior in the dataset. 
    % (b) Our proposed multi-policy planner, which forces the model to explicitly think about multiple policies and choose the most suitable policy given the current driving context.
    The planning trajectories sampled around the initial guess are encoded as the query. The encoded map representation is reshaped into D embeddings with the same dimension as the query. 
    The multi-head cross-attention module then outputs 2-dimensional vectors representing the recommended velocity at each query point, effectively guiding the trajectory optimization process.
  }
  \label{fig:cross attention}
\end{figure}
Implicit \textit{Velocity Field} is a module that maps position $(x, y)$ and time stamp $t$ queries to advised velocity vectors $\bm{v} = f\left( x, y, t \vert \mathcal{S}\right)$ regarding the scenario context embedding. The attention mechanism computes the distance between query~(Q) and key~(K) as a weight, and multiplies it with the value~(V), as depicted in Fig.\ref{fig:cross attention}. This approach effectively captures the relationships and interactions among various elements in the driving context.
% Mathematically expressed as (\ref{eq:attention})
% \begin{equation}
%     \label{eq:attention}
%     \text{Attention}(Q,K,V) = \text{softmax}(\frac{QK^\top}{\sqrt{d_k}})V \notag
% \end{equation}

% This approach can intuitively propagate spatiotemporal distance measurements to represent driving feasibility. 
We incorporate the querying framework into our method. Positions with time embedding $(x, y, t, \sin(t), \cos(t))$ serve as queries for velocity scale in orthogonal directions $(\dot{x}, \dot{y})$. 
Linear layers are implemented to embed tokens and generate velocity vectors before and after the attention module, effectively capturing the dynamics of the driving context and generating appropriate guidance for trajectory planning.

% Sampler Algorithm pipeline?

The velocity vectors are generated regarding to the trajectory samples $\bm{\mathcal{T}}_{sample} = [\mathcal{T}_1,...,\mathcal{T}_N]$ by equation
\begin{equation}
    \mathcal{V}^{t,n}_{p,M} = \mathcal{M}(\bm{p}^{t,n}_\mathcal{T} \vert \mathcal{S}), \notag
\end{equation}
where $\mathcal{M}$ is the velocity inquiry function, $p^{t,n}_\mathcal{T}$ is the position of number $n$ trajectory point at time $t$, and $\mathcal{V}^{t,n}_{p,M}$ is the velocity vector we get from the implicit \textit{Velocity Field} module.
The velocity of the nearest n samples is considered to be an acceptable velocity value. Each velocity imitation loss is multiplied by an exponential discount regarding L2 distance between $[x,y, yaw]$ of the demonstration trajectory and the sampled location by
\begin{equation}
    \mathcal{L}_{imit} = \frac{1}{Z}\sum^{N}_{n=0}\sum^{T}_{t=0}{e^{-\frac{\lVert \bm{p}^{t,n}_{\mathcal{T},yaw}-\bm{p}^t_{\mathcal{T}_{e},yaw}\rVert_2}{2}}(\mathcal{V}^{t,n}_{p,M}-\mathcal{V}^{t,n}_{p,\mathcal{T}})}. \notag
\end{equation}
We introduce a velocity correction loss to assign the appropriate velocity to each position. It is assumed that the velocity at the current time step should be capable of guiding the ego vehicle along the demonstration trajectory point in the subsequent time step. This design is to ensure the planning remains consistent with the desired trajectory while maintaining smoothness and safeness. The correction loss is defined by
\begin{align}
    \hat{\mathcal{V}}^{t,n}_{p,\mathcal{T}} & = \left(\bm{p}^{t+1,n}_{\mathcal{T}_e}-\bm{p}^{t,n}_{\mathcal{T}}\right)/dt, \notag \\
    \mathcal{L}_{correct} & = \frac{1}{Z}\sum^{N}_{n=0}\sum^{T-1}_{t=0}{e^{-\frac{\lVert \bm{p}^{t,n}_{\mathcal{T}}-\bm{p}^t_{\mathcal{T}_{e}}\rVert_2}{2}} \lVert\hat{\mathcal{V}}^{t,n}_{p,\mathcal{T}} - \mathcal{V}^{t,n}_{p,M} \rVert_2}. \notag
\end{align}
The \textit{Velocity Field} construction loss is defined as
\begin{equation}
    \mathcal{L}_{VF} = \mathcal{L}_{imit} + \mathcal{L}_{correct}. \notag
\end{equation}

% DETR
% attention 机制对相似性的构建，使其构建map的合理性
% 构建过程中的loss设计
% 
% \subsection{Measurement of Driving Context}
% The boundary of the ego vehicle is presented by a multi-circle model according to their lengths. All the following distances are calculated by center Cartesian distance minus the nearest pairs of circle radius. 
%TODO demo img
% Three Euler distances are measured, the distance to the closest reference lane $d_{ref}$, the distance to the traffic light $d_{tl}$, and the distance to static traffic obstacles $d_{sdf}$. The raw measurements are denoted as $\mathcal{D}  = [d_{ref},d_{sdf},d_{tl}]$. The elements are mapped to the value space of the signed risk map by
% \begin{equation}
%   \label{eq:maprisk}
%   \mathcal{R}_\mathcal{D} = \beta \ast \exp\left( \lambda \ast \mathcal{D} \right),
% \end{equation}
% where $\beta$ and $\lambda$ are generated by an MLP regarding each encoded context feature.
% The probability of collision with dynamic vehicles of each sample at time step $t$ is 
% \begin{equation}
%   \label{eq:colli}
%   \begin{split}
%     \mathcal{R}_{col} = \mathcal{P}_{t}, %\sum_{a = 1}^{N_a}\sum_{m = 1}^{M}\mathcal{P}_{amt},
%   \end{split}
% \end{equation}
% where $f$ represents probability calculation process in equation (\ref{eq: mvn}).

% The gathered four dim probabilistic feature on each sampled trajectory point $\mathcal{R} = \{\mathcal{R}_\mathcal{D}, \mathcal{R}_{col}\}$ is named \textit{RiskMap}. 
% \subsection{Driving Cost Estimation}
\subsection{Traveling Cost Estimation}
The traveling cost estimation function is constructed with Lagrangian items, including acceleration, jerk, steering, steering change, and the velocity difference $\mathcal{V}_{diff}$ between the sample velocity and the map-suggested velocities. 
These components are combined using a weighted sum with learnable coefficients $\bm{w}_c$, enabling the optimization process to efficiently adapt and respond to various driving scenarios. 
The raw measurements vector is 
\begin{equation}
    \mathcal{D} = \sum^{T}_{t=0}{[\ddot{x_t}^2, \dddot{x_t}^2, \dot{H_t}^2, \ddot{H_t}^2, \mathcal{V}_{diff}^2]}, \notag
\end{equation}
and target function can be mathematically represented as
\begin{equation}
    \mathcal{C}_{\mathcal{T}} = \mathcal{D} \bm{w}_c^{\top}. \notag
\end{equation}

% % Including the velocity filter $w_{v}$ that balances the velocity influence to risk value $ R = R_m \ast v \ast w_{v}$. 
% Symbol $w_{smooth}$ represents that preference of driving smoothness $C_{smooth} = (a,s) \ast w_{smooth}$. 
% Second-order smoothness parameters, including acceleration $a$ and steering $s$, represent yaw rate differed by ground truth and sampled trajectory. 
% The sampled trajectories are generated by a lattice planner \cite{optimalTraj,lattice}.
% Difference between expected target velocity $\bar{v}$ and current velocity $ v$
% $
%   d_v = \bar{v} - v  
% $
% is also taken into consideration, where $\bar{v}$ is generated by a swallow decoder.
% The measurement feature vector is concatenated together. The cost of each trajectory is 
% \begin{equation}
%   \mathcal{C}_{\mathcal{S}} = cat(\mathcal{R},C_{smooth},d_v \times w_d),
% %   \mathcal{C}_{\mathcal{P}i}  = M \ast w_{a}.
% \end{equation}
% where $w_d$ is a coefficient of velocity difference learned by the model.

Different from NMP\cite{NMP}, which directly maps Average Distance Error~(ADE) between expert trajectory and negative samples to traveling cost to smooth the cost value of negative samples, we affine the average distance to probabilistic space.
% Excluding minimizing the driving cost, expected velocity $\bar{v}$ is individually regressed by 
% \begin{equation}
%   \mathcal{L}_v = \sum_{t = 0}^{T} (\bar{v}^t - v^{t}_{\mathcal{T}_{demo}})^2,
% \end{equation}
% regulating items that are used to avoid over-fitting is introduced. 
The selection loss calculates the Cross Entropy~(CE) of cost distribution and distance distribution of sampling trajectory set $\mathcal{T}_s $ by 
\begin{align}
  \mathcal{P}(\text{ADE}) & = \text{Softmax}\left( 1- \eta \left( \lVert \bm{\mathcal{T}}_{s} - \mathcal{T}_{e} \rVert_2\right)\right), \notag \\
  \mathcal{P}(\mathcal{C}_{\bm{\mathcal{T}}_{samp}}) & = \text{Softmax}\left(1-\eta \left( \mathcal{C}_{\bm{\mathcal{T}}_{samp}} \right)\right), \notag \\
  \mathcal{L}_{sele} & = \text{CE}\left(\mathcal{P}(\mathcal{C}_{\bm{\mathcal{T}}_{s}}),\mathcal{P}(\text{ADE})\right), \notag
\end{align} 
where $\eta$ is the normalization function along sampled candidate trajectories.
Please note that not all the samples are taken into loss calculation, only 20 samples with lower cost are considered.

In addition to $\mathcal{L}_{VF}$ and $\mathcal{L}_{sele}$, we introduce a multi-modal Imitation Learning planning and multi-agent prediction head to generate initial guesses and provide multi-modality information, respectively. For the initial guessing and prediction head, the loss is defined as follows:
\begin{equation}
    \mathcal{L}_{IL} = \mathcal{L}_{ADE}^{plan} + \mathcal{L}_{FDE}^{plan} + \mathcal{L}_{ADE}^{pre} + \mathcal{L}_{modal}, \notag
\end{equation}
where $\mathcal{L}_{modal}$ is the cross-entropy loss of modal selection. The modal with the minimum summation of the plan and prediction error is expected to have the highest probability.
Finally, the overall training loss is  
\begin{equation}
  \mathcal{L} = \mathcal{L}_{VF} + \mathcal{L}_{sele} + \mathcal{L}_{IL}. \notag
\end{equation}

% Table 1
\section{Experiment}

\begin{table*}[t]
  \centering
\caption{Open-loop Comparison}
\label{tab:openloop}
\renewcommand\tabcolsep{10.0pt}
\begin{tabular}{@{}lcccccccccc@{}}
\toprule
Methods  & Collision$\downarrow$       & \begin{tabular}[c]{@{}c@{}}Off\\ Route$\downarrow$ \end{tabular}     & \begin{tabular}[c]{@{}c@{}}Traffic\\ Light$\downarrow$ \end{tabular} & Acc$\downarrow$              & Jerk$\downarrow$             & \begin{tabular}[c]{@{}c@{}}Prediction \\ ADE$\downarrow$ \end{tabular} & \begin{tabular}[c]{@{}c@{}}Prediction \\ FDE$\downarrow$ \end{tabular} & \begin{tabular}[c]{@{}c@{}}Plan\\ L2@1s$\downarrow$ \end{tabular} & \begin{tabular}[c]{@{}c@{}}Plan\\ L2@3s$\downarrow$ \end{tabular} & \begin{tabular}[c]{@{}c@{}}Plan\\ L2@5s$\downarrow$ \end{tabular} \\ \midrule
Human    & 0.12\%          & 0.70\%                                              & 2.38\%                                                  & 0.6546          & 3.2496          & 0.0000                                                    & 0.0000                                                    & 0.0000                                               & 0.0000                                               & 0.0000                                               \\ \midrule
IL       & \textbf{1.44\%} & 1.26\%                                              & 1.44\%                                                  & 0.5794          & 2.5236          & 0.6893                                                    & \textbf{1.7352}                                           & \textbf{0.0920}                                               & \textbf{0.7191}                                               & \textbf{2.1360}                                               \\
% IL(DA) & 11.15\%         & 31.73\%                                             & 1.49\%                                                  & 1.7311          & 7.9794          & 0.7096                                                    & 1.7803                                                    & 0.2527                                               & 0.8865                                               & \textbf{2.3331}                                      \\
DIPP\cite{DIPP}     & 6.70\%          & 5.08\%                                              & 1.49\%                                                  & 2.4164          & 27.9553         & \textbf{0.6547}                                           & 3.2497                                                    & 0.3144                                               & 1.4119                                               & 3.5733                                               \\
% EULA   & 2.28\%          & 1.67\%                                              & \textbf{1.42\%}                                         & 0.4631          & 0.2386          & 0.7115                                                    & 1.7834                                                    & 0.1415                                               & 0.7600                                               & 2.2181                                               \\
% CF     &                 &                                                         &                                                     &                 &                 &                                                           &                                                           &                                                      & \textbf{}                                            &                                                      \\
% VF     & \textbf{1.22\%} & \textbf{1.26\%}                                     & 1.51\%                                                  & 0.4712          & \textbf{0.2224} & 0.7133                                                    & 1.7457                                                    & \textbf{0.0773}                                      & \textbf{0.6990}                                      & \textbf{2.0701}                                      \\ \midrule
EULA     & 9.65\%          & 1.69\%                                              & \textbf{1.36\%}                                         & \textbf{0.4470} & 0.3350          & 0.7406                                                    & 1.8617                                                    & 0.2305                                               & 1.1124                                               & 2.7648                                               \\
CF       & 6.26\%          & \textbf{0.32\%}                                              & 1.61\%                                                  & 0.8026          & 4.5039          & 0.7379                                                    & 1.8579                                                    & 0.4014                                               & 3.0564                                               & 8.2568                                               \\
VF(ours)       & 8.01\%          & 1.65\%                                     & 1.41\%                                                  & 0.4850          & \textbf{0.2464} & 0.7071                                          & 1.7555                                           & 0.1371                                      & 0.8743                                     & 2.4642                                               \\ \bottomrule
\end{tabular}
\end{table*}

\begin{table*}[t]
\centering
\caption{Closed-loop Comparison}
\label{tab:closeloop}
\renewcommand\tabcolsep{10.0pt}
\begin{tabular}{@{}lcccccccccc@{}}
\toprule
Methods        & Collision$\downarrow$        & \begin{tabular}[c]{@{}c@{}}Off\\ Route$\downarrow$ \end{tabular} & \begin{tabular}[c]{@{}c@{}}Traffic\\ Light$\downarrow$ \end{tabular} & Progress$\uparrow$          & Acc$\downarrow$              & Jerk$\downarrow$             & Lat\_Acc$\downarrow$         & \begin{tabular}[c]{@{}c@{}}Plan\\ L2@3s$\downarrow$ \end{tabular} & \begin{tabular}[c]{@{}c@{}}Plan\\ L2@5s$\downarrow$ \end{tabular} & \begin{tabular}[c]{@{}c@{}}Plan\\ L2@10s$\downarrow$ \end{tabular} \\ \midrule
% Human          & 0.50\%          & 1.50\%                                              & 0.50\%                                                  & 77.6694          & 0.6947          & 3.4178          & 0.0883          & 0.0000                                               & 0.0000                                               & 0.0510                                                \\ \midrule
IL             & 40.00\%         & 15.00\%                                             & 1.00\%                                                  & 28.9308          & 1.7357          & 3.5291          & 0.2030          & 5.9404                                               & 10.7390                                              & 25.5545                                               \\
DIPP           & 9.00\%          & 0.00\%                                     & 5.00\%                                                  & 69.7017          & 0.8945          & \textbf{3.4976} & 1.0862          & 2.4340                                               & 4.9987                                               & 11.0548                                               \\
EULA           & 11.50\%         & 4.00\%                                              & 3.00\%                                                  & \textbf{78.3942} & 0.7345          & 5.6692          & 0.0961          & 2.3328                                               & 4.5252                                               & 9.4676                                                \\
CF     & 21.00\%         & 1.00\%                                              & \textbf{0.00\%}                                                  & 49.0182          & \textbf{0.5945} & 4.1452          & \textbf{0.0650} & 2.1357                                               & 4.5088                                               & 10.8249                                               \\
VF & \textbf{6.00\%} & \textbf{0.00\%}                                              & 1.00\%                                       & 55.0490          & 0.8580          & 7.7434          & 0.1101          & \textbf{1.9397}                                      & \textbf{3.4757}                                      & \textbf{8.9564}                                       \\ \bottomrule
\end{tabular}
\end{table*}

\subsection{Implementation details}

\textit{1) Dataset:}
Our experiments are conducted on the Waymo open motion Dataset \cite{waymo_dataset}, which contains 104k, 20s time horizon, and real-world driving samples, collected on a complex urban route.
HD maps and annotations for traffic signals/participants are provided at 10 Hz.
1000 packages with ~100 scenarios in each package are provided. 
The 0 to 699 and 700 to 899 are segmented as training and validating datasets respectively. The scenarios are processed into 7-second object tracks. 
The training set consists 641943 tracks and validating set consists 182731 tracks. 
For each track, the first 2 seconds are considered as track history and the following 5 seconds are expert planning track and prediction ground truth. 
The first scenarios in files 900 to 999 are for closed-loop testing and all the scenarios in those files are for open-loop testing.
% A subset of about 3.6M frames is randomly selected for training and tested on a subset of the original validation set which contains 3600 frames.

\begin{equation}
    \label{eq:u_aug}
    \mathcal{U}_{his}^{aug} = \mathcal{U}_{his} + \mathcal{N}(0,\mathcal{\varepsilon}\cdot\mathbf{u}_{lim})
\end{equation}
\begin{equation}
    \label{eq:X_aug}
    \mathcal{X}_{his}^{aug} = BicycleModel(\mathcal{U}_{his}^{aug}, \mathcal{X}_{his}^{t=0})
\end{equation}

To reduce the domain shift problem in imitation learning, we propose a novel approach that leverages a time-dependent Gaussian distribution to augment the control variables of the driving history. 
Specifically, we apply equations (\ref{eq:u_aug}) and (\ref{eq:X_aug}) to introduce variability to the control inputs, while leaving the future trajectory unchanged. Data augmentation is implemented in all the experiments without special statements.
% The model will automatically generate a trackable trajectory minimizing L2 distance.

\textit{2) Training:}
Training with AdamW\cite{adamw} optimizer and learning rate 1e-4 for 20 epochs are conducted on a server equipped with $4\times$RTX3090 GPU and $2\times48$ threads Intel(R) Xeon(R) Gold 5318S CPU @2.10GHz.
The iterative sampling optimizer conducts 3 iterations in the training process and 10 iterations in validating process. 
The best 10 samples are selected as parents in the iterative sampling process. 
% \subsection{Data Augmentation}
% \subsection{Training and Inference Details} \label{sec:aug}
A two-stage supervised learning process is conducted. We first train the prediction model from all the vehicles for 2 epochs. Then the map parameter and cost function training process is conducted by imitating the behavior demo of the ego vehicle.
% The raw context information is processed the same as in the first stage, and prediction results are inputs of the second stage.

\textit{3) Evaluation:}
We compare our model and other baselines on the dataset with common metrics. Average Distance Error(ADE)$(m)$, and Final Distance Error(FDE)$(m)$ are introduced to illustrate the ability to \textbf{imitate} expert behavior. Collision Rate(CR), Off-Road Rate(ORR), and Traffic Light Violation(TLV), are \textbf{safety} measurements. Acceleration (Acc)$(m/s^2)$, Jerk$(m/s^3)$ measures the \textbf{comfortness}.  L2 distance at typical time steps $(L2@Ns)$ is introduced to reveal more accurate distance errors.

% While it is known that open-loop evaluation does not reflect the actual performance of the model for planning problems as imitation learning suffers from covariant shift \cite{covariate_shift}, we include this for completeness and consider it as an auxiliary evaluation.

\subsection{Baselines}
% TODO three images comparing sample cost distribution

% Baselines are developed to illustrate the profiles of our method.
% We compare the planning results against two representative baselines on the Lyft dataset.
% \textbf{\textit{Imitation Learning (IL)}}, where the model share the same encoder with our MPNP but only outputs a single modal trajectory.
% \textbf{\textit{Imitation Learning with multi-modal outputs (IL-MM)}}, which is the same as MPNP except that the multi-modal outputs are not conditioned on reference lines.
% IL-MM is trained with the winner-takes-all strategy.
\textbf{\textit{Imitation Learning~(IL)}}, where the input is vectorized context\cite{vectornet} and directly outputs multi-modal future ego plan controls and trajectory of other agents.
% \textbf{\textit{Imitation Learning with data augmentation}}

\textbf{\textit{Differentiable Integrated Motion Prediction and Planning~(DIPP)}} is inverse optimal control based planning method. Where the IL model is implemented to offer initial guess and trajectory prediction of other agents. Euclidean distance to on-road objects and trajectory smoothness are weighted sums to estimate driving costs\cite{DIPP}. Please note that data augmentation is not applicable to this method.

\textbf{\textit{Sampling Based DIPP~(EULA)}} also measures Euclidean distances using the same cost function as DIPP but implemented with our proposed iterative sampling planner in both the training and evaluating processes.

\textbf{\textit{Cost Field~(CF)}} use the same framework as our proposed Velocity Field, but the vector outputs are summed to be scalar to directly represent the driving cost of the inquired point. 

\textbf{\textit{Velocity Field~(VF)}} our proposed method, the implementing details are described in Section \ref{sec:method}.

% \subsection{Qualitative comparison}
% Fig shows the distribution of driving costs of EULA, CF and VF
% Qualitative comparison among those methods is shown in %\ref{fig: gmm}.
% TODO: comparison images

% [explaination]

\subsection{Open-loop Comparison}

% \usepackage{booktabs}

Table \ref{tab:openloop} presents the open-loop results of the methods described above. 
The first row labeled \textbf{Human} indicates that the data in the simulation system has some noise and inaccuracies, despite being roughly accurate. 
In general, all deep neural network-based methods, including IL, are able to accurately imitate the behavior of human drivers, demonstrating the effectiveness of deep learning in pattern recognition. 
Comparing DIPP and EULA, we observe a reduction in imitation metrics, indicating that our iterative sampling planner can find low-cost trajectories that closely resemble the expert demonstrations. 
By comparing the CF and VF metrics, we find that incorporating the prior velocity vector can improve the accuracy of the cost calculation and lead to more precise results. 
Overall, these open-loop results suggest that our proposed method can achieve comparable or better performance than the state-of-the-art baselines in terms of imitating human driving behavior, safeness, and comfortness.

\subsection{Closed-loop Comparison}

Table \ref{tab:closeloop} presents the closed-loop results, which provide a more realistic assessment of the model's ability to handle domain shift and generalize to real-world driving scenarios. 
Despite the use of data augmentation, the deep neural network (DNN) planner struggles to cope with complex dynamic driving environments. 
In contrast, DIPP leverages manually designed rules to optimize the planning strategy of the model, resulting in significant improvements in planning security. 
By comparing DIPP, and EULA, we found that the planning module used in EULA is not as effective as Gauss-Newton for finding the optimal trajectory. 
But it still outperforms the DNN-based methods in terms of safety and similarity to human drivers and achieves comparable results as DIPP. 
For the CF method, the worse result in collision rate, progress and similarity compared with VF proves its limited capability. Because it can only provide first-order information for the downstream optimization module.
The VF method effectively uses expert data to form an information-rich local map, avoiding inductive bias due to manual design and various traffic information interaction modes. 
As a result, VF achieves better results in terms of safety and imitation ability.

\subsection{Operation Time}
Operation time evaluation is conducted on a PC equipped with Intel 12900K and NVIDIA RTX3080Ti. The average operation time for generating a planning result is 0.048s(Min. 0.017, Max. 0.111). The operating time meets the real-time requirement.
\section{Conclusion}

% We have presented a novel multi-policy neural planner (MPNP) for urban driving. 
% The core idea is to explicitly generate trajectories conditioned on different polices, which are represented in the form of reference lines. 
% We also analyzed the causal confusion problem, and proposed an effective augmentation method addressing it. 
% MPNP exhibits the richest policy diversity among all baselines, and achieves competitive performance in success rate. 
% It also shows the ability to capture diverse driving behaviors beyond lane-keeping from very limited demonstration, while other SOTA methods fail to. 

In this work, we proved that \textit{Velocity Field} is an informative way to offer traveling costs for local path planning,
% an efficient multi-modal prediction result utilization method for local planning by unifying it together with other driving contexts and diagnosing it as Velocity Field.
and demonstrate the effectiveness of the iterative sampling planner to generate safe trajectories by experiments. 
Our method significantly improves the planning performance compared to the baseline DIPP method. 
Specifically, it reduces the probability of collision by 33.3\%, the probability of running a red light by 80\%, and achieves the highest improvement of 43.81\% in similarity with human drivers. 
% The velocity field can effectively provide strategy priors to the downstream planner. 
Overall, our approach offers a promising solution for improving the planning performance of autonomous driving systems and enhancing their ability to imitate human driving behavior.

\bibliographystyle{IEEEtran}
\bibliography{IEEEabrv,ref}

\begin{thebibliography}{10}
\providecommand{\url}[1]{#1}
\csname url@rmstyle\endcsname
\providecommand{\newblock}{\relax}
\providecommand{\bibinfo}[2]{#2}
\providecommand\BIBentrySTDinterwordspacing{\spaceskip=0pt\relax}
\providecommand\BIBentryALTinterwordstretchfactor{4}
\providecommand\BIBentryALTinterwordspacing{\spaceskip=\fontdimen2\font plus
\BIBentryALTinterwordstretchfactor\fontdimen3\font minus \fontdimen4\font\relax}
\providecommand\BIBforeignlanguage[2]{{%
\expandafter\ifx\csname l@#1\endcsname\relax
\typeout{** WARNING: IEEEtran.bst: No hyphenation pattern has been}%
\typeout{** loaded for the language `#1'. Using the pattern for}%
\typeout{** the default language instead.}%
\else
\language=\csname l@#1\endcsname
\fi
#2}}

\bibitem{urban_driver}
O.~Scheel, L.~Bergamini, M.~Wolczyk, B.~Osi{\'n}ski, and P.~Ondruska, ``Urban driver: Learning to drive from real-world demonstrations using policy gradients,'' in \emph{Conf. Rob. Learing}.\hskip 1em plus 0.5em minus 0.4em\relax PMLR, 2022, pp. 718--728.

\bibitem{NMP}
W.~Zeng, W.~Luo, S.~Suo, A.~Sadat, B.~Yang, S.~Casas, and R.~Urtasun, ``End-to-end interpretable neural motion planner,'' in \emph{Proc. of the IEEE/CVF Conference on Computer Vision and Pattern Recognition}, 2019, pp. 8660--8669.

\bibitem{attention}
A.~Vaswani, N.~Shazeer, N.~Parmar, J.~Uszkoreit, L.~Jones, A.~N. Gomez, {\L}.~Kaiser, and I.~Polosukhin, ``Attention is all you need,'' \emph{Advances in neural information processing systems}, vol.~30, 2017.

\bibitem{OCC_Map}
H.~Moravec and A.~Elfes, ``High resolution maps from wide angle sonar,'' in \emph{Proceedings. 1985 IEEE International Conference on Robotics and Automation}, vol.~2, 1985, pp. 116--121.

\bibitem{occflow}
R.~Mahjourian, J.~Kim, Y.~Chai, M.~Tan, B.~Sapp, and D.~Anguelov, ``Occupancy flow fields for motion forecasting in autonomous driving,'' \emph{IEEE Robotics and Automation Letters}, vol.~7, no.~2, pp. 5639--5646, 2022.

\bibitem{CoverNet}
\BIBentryALTinterwordspacing
T.~Phan{-}Minh, E.~C. Grigore, F.~A. Boulton, O.~Beijbom, and E.~M. Wolff, ``Covernet: Multimodal behavior prediction using trajectory sets,'' \emph{CoRR}, vol. abs/1911.10298, 2019. [Online]. Available: \url{http://arxiv.org/abs/1911.10298}
\BIBentrySTDinterwordspacing

\bibitem{Chauffeurnet}
M.~Bansal, A.~Krizhevsky, and A.~Ogale, ``Chauffeur{N}et: Learning to drive by imitating the best and synthesizing the worst,'' in \emph{Rob. Sci. Systems}, FreiburgimBreisgau, Germany, June 2019.

\bibitem{vectornet}
J.~Gao, C.~Sun, H.~Zhao, Y.~Shen, D.~Anguelov, C.~Li, and C.~Schmid, ``Vectornet: Encoding {HD} maps and agent dynamics from vectorized representation,'' in \emph{Proc. of the IEEE/CVF Conference on Computer Vision and Pattern Recognition}, 2020, pp. 11\,525--11\,533.

\bibitem{PGP}
N.~Deo, E.~Wolff, and O.~Beijbom, ``Multimodal trajectory prediction conditioned on lane-graph traversals,'' in \emph{5th Annual Conference on Robot Learning}, 2021.

\bibitem{cj2022mpnp}
J.~Cheng, R.~Xin, S.~Wang, and M.~Liu, ``Mpnp: Multi-policy neural planner for urban driving,'' in \emph{{IEEE/RSJ} Int. Conf. Intell. Rob. Sys. (IROS)}.\hskip 1em plus 0.5em minus 0.4em\relax IEEE, 2022.

\bibitem{hdmap}
R.~Liu, J.~Wang, and B.~Zhang, ``High definition map for automated driving: Overview and analysis,'' \emph{The Journal of Navigation}, vol.~73, no.~2, pp. 324--341, 2020.

\bibitem{learn2predict}
H.~Song, D.~Luan, W.~Ding, M.~Y. Wang, and Q.~Chen, ``Learning to predict vehicle trajectories with model-based planning,'' in \emph{5th Annual Conference on Robot Learning}, 2021.

\bibitem{unlikelihood}
\BIBentryALTinterwordspacing
D.~Zhu, M.~Zahran, L.~E. Li, and M.~Elhoseiny, ``Motion forecasting with unlikelihood training in continuous space,'' in \emph{Proceedings of the 5th Conference on Robot Learning}, ser. Proceedings of Machine Learning Research, A.~Faust, D.~Hsu, and G.~Neumann, Eds., vol. 164.\hskip 1em plus 0.5em minus 0.4em\relax PMLR, 08--11 Nov 2022, pp. 1003--1012. [Online]. Available: \url{https://proceedings.mlr.press/v164/zhu22a.html}
\BIBentrySTDinterwordspacing

\bibitem{MP3}
S.~Casas, A.~Sadat, and R.~Urtasun, ``Mp3: A unified model to map, perceive, predict and plan,'' in \emph{Proc. of the IEEE/CVF Conference on Computer Vision and Pattern Recognition}, 2021, pp. 14\,403--14\,412.

\bibitem{Nelson2007VFPF}
D.~R. Nelson, D.~B. Barber, T.~W. McLain, and R.~W. Beard, ``Vector field path following for miniature air vehicles,'' \emph{IEEE Transactions on Robotics}, vol.~23, no.~3, pp. 519--529, 2007.

\bibitem{lawrence2008lyapunov}
D.~A. Lawrence, E.~W. Frew, and W.~J. Pisano, ``Lyapunov vector fields for autonomous unmanned aircraft flight control,'' \emph{Journal of Guidance, Control, and Dynamics}, vol.~31, no.~5, pp. 1220--1229, 2008.

\bibitem{agro2023implicit}
B.~Agro, Q.~Sykora, S.~Casas, and R.~Urtasun, ``Implicit occupancy flow fields for perception and prediction in self-driving,'' in \emph{Proceedings of the IEEE/CVF Conference on Computer Vision and Pattern Recognition}, 2023, pp. 1379--1388.

\bibitem{yaman2013survey}
F.~Yaman, V.~G. Yakhno, and R.~Potthast, ``A survey on inverse problems for applied sciences,'' \emph{Mathematical problems in engineering}, vol. 2013, 2013.

\bibitem{levine_continuous}
S.~Levine and V.~Koltun, ``Continuous inverse optimal control with locally optimal examples,'' \emph{arXiv preprint arXiv:1206.4617}, 2012.

\bibitem{DIPP}
Z.~Huang, H.~Liu, J.~Wu, and C.~Lv, ``Differentiable integrated motion prediction and planning with learnable cost function for autonomous driving,'' \emph{arXiv preprint arXiv:2207.10422}, 2022.

\bibitem{STOMP}
M.~Kalakrishnan, S.~Chitta, E.~Theodorou, P.~Pastor, and S.~Schaal, ``Stomp: Stochastic trajectory optimization for motion planning,'' in \emph{2011 IEEE International Conference on Robotics and Automation}, 2011, pp. 4569--4574.

\bibitem{lattice}
M.~McNaughton, C.~Urmson, J.~M. Dolan, and J.-W. Lee, ``Motion planning for autonomous driving with a conformal spatiotemporal lattice,'' in \emph{2011 IEEE International Conference on Robotics and Automation}.\hskip 1em plus 0.5em minus 0.4em\relax IEEE, 2011, pp. 4889--4895.

\bibitem{waymo_dataset}
S.~Ettinger, S.~Cheng, B.~Caine, C.~Liu, H.~Zhao, S.~Pradhan, Y.~Chai, B.~Sapp, C.~R. Qi, Y.~Zhou, \emph{et~al.}, ``Large scale interactive motion forecasting for autonomous driving: The waymo open motion dataset,'' in \emph{Proc. {IEEE/CVF} Conf. Comput. Vis. Parttern Recognit. (CVPR)}, 2021, pp. 9710--9719.

\bibitem{adamw}
I.~Loshchilov and F.~Hutter, ``Decoupled weight decay regularization,'' in \emph{International Conference on Learning Representations}, 2018.

\end{thebibliography}

\end{document}